\documentclass[lettersize,journal]{IEEEtran}
\usepackage{amsmath,amsfonts}
\usepackage{algorithmic}
\usepackage{algorithm}
\usepackage{array}
\usepackage[caption=false,font=normalsize,labelfont=sf,textfont=sf]{subfig}
\usepackage{textcomp}
\usepackage{stfloats}
\usepackage{url}
\usepackage{verbatim}
\usepackage{graphicx}
\usepackage{cite}
\usepackage{multicol}
\usepackage{multirow}
\usepackage{booktabs}
\usepackage{hyperref}
\begin{document}

\title{CARNet: Collaborative Adversarial Resilience for Robust Underwater Image Enhancement and Perception}

\author{Zengxi~Zhang,~Zeru~Shi,~Zhiying~Jiang~and~Jinyuan~Liu,~\IEEEmembership{Member,~IEEE}
	\thanks{This work was supported in part by the National Natural Science Foundation of China (Nos. 62302078, 62372080); and in part by China Postdoctoral Science Foundation (No. 2023M730741).
	\\\emph{(Corresponding author: Zhiying Jiang.)}}
	\thanks{Zengxi Zhang and Zeru Shi are with the School of Software Technology, Dalian University of Technology, Dalian 116024, China (e-mail: cyouzoukyuu@gmail.com; shizeru77@gmail.com).}
	\thanks{Zhiying Jiang is with the College of Information Science and Technology, Dalian Maritime University, Dalian 116026, China (e-mail: zyjiang0630@gmail.com).}
	\thanks{Jinyuan Liu is with the School of Mechanical Engineering, Dalian University of Technology, Dalian 116024, China (e-mail: atlantis918@hotmail.com).}
}

\markboth{Journal of \LaTeX\ Class Files,~Vol.~14, No.~8, August~2021}%
{Shell \MakeLowercase{\textit{et al.}}: A Sample Article Using IEEEtran.cls for IEEE Journals}

\IEEEpubid{0000--0000/00\$00.00~\copyright~2021 IEEE}

\maketitle
\begin{figure*}[!h]
	\centering
	\setlength{\tabcolsep}{1pt}
	\includegraphics[width=0.95\textwidth]{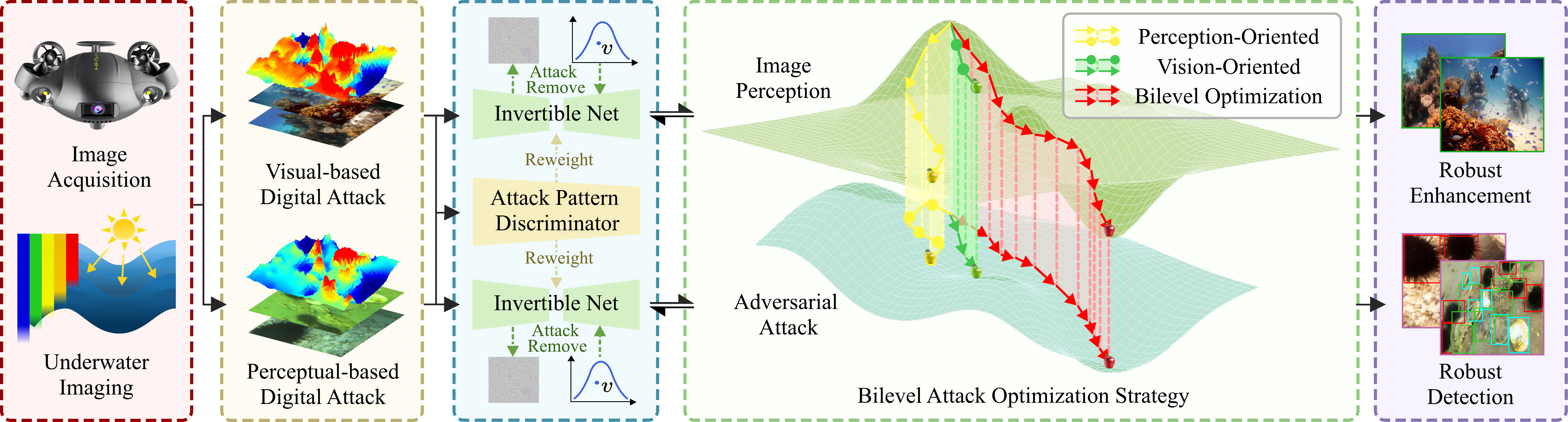}
	\caption{Overview of the underwater adversarial attack removal process. The proposed method can adaptively adjust the parameter weights of the network by discriminating the different attack patterns. Under the guidance of bilevel attack optimization strategy, the proposed method can effectively improve the effect of underwater image perception under attacks.}
	\label{fig:firstfigure}
\end{figure*}
\begin{abstract}
	Due to the uneven absorption of different light wavelengths in aquatic environments, underwater images suffer from low visibility and clear color deviations. With the advancement of autonomous underwater vehicles, extensive research has been conducted on learning-based underwater enhancement algorithms. These works can generate visually pleasing enhanced images and mitigate the adverse effects of degraded images on subsequent perception tasks. However, learning-based methods are susceptible to the inherent fragility of adversarial attacks, causing significant disruption in enhanced results. In this work, we introduce a collaborative adversarial resilience network, dubbed CARNet, for underwater image enhancement and subsequent detection tasks. 
	Concretely, we first introduce an invertible network with strong perturbation-perceptual abilities to isolate attacks from underwater images, preventing interference with visual quality enhancement and perceptual tasks. 
	Furthermore, an attack pattern discriminator is introduced to adaptively identify and eliminate various types of attacks. 
	Additionally, we propose a bilevel attack optimization strategy to heighten the robustness of the network against different types of attacks under the collaborative adversarial training of vision-driven and perception-driven attacks. 
	Extensive experiments demonstrate that the proposed method outputs visually appealing enhancement images and performs an average 6.71\% higher detection mAP than state-of-the-art methods.
\end{abstract}


\section{Introduction}
\IEEEPARstart{T}{he} marine environment contains a wealth of energy and biological resources, which has garnered considerable attention towards underwater exploration in recent years.
However, due to the refraction of light and the reflection of fine particles underwater, wavelength-dependent light suffers from varying degrees of attenuation even at consistent depths, which significantly affects the visual effect of the image. Such distortions also detract from subsequent detection tasks. 
Therefore, underwater image enhancement has become one of the important computer vision tasks. 
With the development of Autonomous Underwater Vehicles~(AUVs)~\cite{shou2021virtual}, a growing body of research~\cite{jiang2023perception,li2023ruiesr,liu2024underwater} has employed deep learning to enhance underwater images in recent years. Compared to optimization-based methods~\cite{abunaser2015underwater,sethi2015optimal}, deep learning approaches have explored various network architectures and achieved remarkable performance by training on large datasets. Among them, many~\cite{liu2022tacl,jiang2022topal} also have utilized enhancement networks as preprocessing modules for detection tasks to improve prediction performance. 

Despite the advancement of existing underwater image enhancement algorithms, the robustness of underwater enhancement networks has yet to be thoroughly studied. Even slight perturbations can significantly impact the prediction performance of the network. 
We argue that existing underwater image perception networks rely on complex pre-trained models~\cite{jiang2022topal} and multi-scale feature extraction techniques~\cite{li2021ucolor}.
However, these methods usually overlook the potential threats posed by adversarial attacks, which consequently leads to instability and oscillation in the network parameters during the process of adversarial training.

In recent years, adversarial attacks have been applied to many other computer vision tasks~\cite{yin2018deep,zhang2019towards,yu2022towards, gui2023adversarial} to improve the robustness of networks. 
Depending on the specific requirements of various applications, these attacks can be classified into the visually-driven attack~\cite{yin2018deep,gui2023adversarial} and the perception-driven attack~\cite{zhang2019towards,dong2022adversarially}: The visually-driven attack can disrupt the visual restoration effects of degraded images, while the perception-driven attack can significantly reduce the prediction performance of perception tasks. 
However, existing works usually only consider defending a single type of adversarial attack due to the obvious differences between these attacks.

In this paper, we consider adversarial attack as imperceptible high-frequency perturbations. Considering the advantages of invertible networks in high-frequency feature extraction, we propose a collaborative adversarial resilience network dubbed CARNet to enhance underwater images which suffer from different types of attacks, thus facilitating pleasing visual and detection results. Specifically, in the forward process, we transform the underwater image into latent components in different frequency domains and separate out the degradation factors from the underwater image. 
Subsequently, in the reverse process, we reconstruct the clean enhanced image by substituting the separated attack components with the latent variable sampled from a distribution prior and obtain robust detection results.
Furthermore, considering the diversity of different types of attacks, we introduce an attack pattern discriminator in the invertible process, enabling the network to adaptively adjust its parameters with different weights based on the type of attack from the input image. 
Additionally, we propose a bilevel attack optimization strategy, thereby realizing a more comprehensive adversarial defense by harmonizing different types of attacks.

In summary, our contributions are as follows:
\begin{itemize}
	\item We introduce adversarial attacks to underwater tasks and propose a robust network for enhancement and detection tasks that counteracts the attack component in latent representations via an invertible framework.
	\item We propose an attack pattern discriminator which learns different groups of convolution kernels and adjusts the weights of the network adaptively according to the input attack type to ensure reliable robustness.
	\item A bilevel attack optimization strategy is introduced so that the proposed network can simultaneously achieve robust underwater image enhancement and detection.
	\item Extensive experiments on multiple underwater datasets demonstrate the effectiveness of our method.
\end{itemize}

\section{RELATED WORK}

\subsection{Underwater Image Enhancement}
Existing traditional underwater enhancement methods can be classified into model-based methods and model-free methods. Model-based methods often incorporate domain knowledge as priors to assist in solving underwater enhancement problems:
Peng et al.~\cite{peng2018darkchannel} established a new unified method to estimate ambient light based on DCP image restoration, and embedded adaptive color correction in the image imaging model. 
Galdran et al.~\cite{galdran2015redchannel} then derived the dark channel into a red channel to restore the contrast of the image by restoring short-wavelength colors.
Model-free methods usually address the light attenuation phenomenon in underwater images by altering the pixel distribution:
Inspired by Retinex, Zhang et al.~\cite{zhang2017retinex} used a combination of bilateral filters and trilateral filters on different channels of underwater images to achieve adaptive enhancement of underwater images.
Inspired by the morphology of the teleost retina, Gao et al.~\cite{gao2019arm} used ganglion cells with color confrontation to correct the color of the image, and proposed a brightness-based fusion strategy to generate fusion-enhanced images based on different paths of the retina.
Zhang et al.~\cite{zhang2022mmle} combined the maximum attenuation map-guided fusion strategy and the minimum color loss principle to guide the correction of image color details. And use the mean and variance calculated from the integral map and the squared integral map to adaptively enhance the image contrast.

There are also many deep learning-based approaches~\cite{liu2022tacl,zhang2023waterflow,huang2023contrastive} to improve the visual quality of underwater image through data-driven networks:
Liu et al.~\cite{liu2022tacl} combined contrastive learning with a bilaterally constrained adversarial strategy to retain the information features of the image through the two-way coupling of the Siamese network.
Zhang et al.~\cite{zhang2023waterflow} proposed a detection-driven normalization flow to achieve a balance between visual enhancement and detection applications through the transmission of high-level semantic information.
Huang et al.~\cite{huang2023contrastive} proposed a semi-supervised image restoration framework based on the average teacher principle, which improves the robustness of the network through domain knowledge of unlabeled images.
Although existing methods can already achieve good enhancement effects, little work has been devoted to studying the vulnerability of underwater enhancement networks to adversarial attacks.
\subsection{Invertible Neural Network}
Invertible neural networks (INNs) enable both forward and backward propagation within the same framework through the reversible mapping between the source domain x and target domain y. In recent years, INNs have been widely applied in computer vision tasks:
Xiao et al.~\cite{xiao2020invertiblesr} proposed a reversible super-resolution network aiming to optimize the processes of upscaling and downscaling synchronously.
Yu et al.~\cite{yu2020efficient} build a bidirectional adaptive encoder and improves the  ability of the network to capture temporal dependencies while reducing memory consumption in video prediction.
Liu et al.~\cite{liu2021invertibledenoise} transformed noisy images into low-resolution and high-frequency high-frequency information containing noise. They then replaced this information with fixed regularized distributions and obtained clean images through the reverse process.
Zhao et al.~\cite{zhao2021invertible} transformed color information into Gaussian distribution through a invertible network to achieve reversible color decolorization and restoration.
Jing et al.~\cite{jing2021hinet} achieved image hiding by synchronously learning the processes of image concealing and revealing, with the secret image encoded into the wavelet domain.
Zhu et al.~\cite{zhu2022bijective} coupled the learning processes of shadow removal and shadow generation in a unified parameter-sharing framework, achieving shadow removal. 
The purpose of this work is to leverage the encoding ability of invertible networks for high-frequency information to remove potential attacks in the input images.
\subsection{Object Detection}
With the advent of deep learning, object detection has experienced rapid advancements in recent years. Existing object detection methods are categorized into two-stage and single-stage detections based on their processing steps. Two-stage methods~\cite{ren2015faster,liu2020cbnet,sun2021sparse} first extract regions of interest from the image and subsequently classify them using a classification neural network. Conversely, single-stage methods~\cite{liu2016ssd,xie2022pyramid} typically employ  a single network to concurrently perform object classification and localization.

Existing detectors usually perform well primarily on clear in-air images. However, underwater images suffer from light degradation and blurring due to fine particles, which distort the semantic information of underwater scenes. To enable detectors to better handle degraded images, existing methods~\cite{liu2022tacl,zhang2023waterflow} have integrated underwater image enhancement as a preprocessing step for object detection to improve performance on degraded images. Despite these efforts, the high-dimensional space managed by deep networks remains highly susceptible to attacks, where even minor perturbations can result in significant misclassifications. Therefore, enhancing the robustness of object detection networks is still a critical area for continued research.

\subsection{Adversarial Attack and Defense}
Despite the outstanding representation capability of deep networks, it can also be damaged by slight perturbations. In recent years, various attack methods have been proposed, which add visually imperceptible perturbations to input images, resulting in inaccurate predictions. The existing attack methods are roughly divided into black box-based attack methods~\cite{sarkar2017upset,yang2020learning} and white box-based attack methods.~\cite{goodfellow2014explaining,kurakin2018adversarial,madry2017pgd}
The main difference between these attacks is the degree of understanding of the network. The proposed method attacks the model through white-box attacks.
Considering the linearity of neural networks, the Fast Gradient Sign Method (FGSM)~\cite{goodfellow2014explaining} is first proposed to generate adversarial examples.
Different from FGSM, I-FGCM~\cite{kurakin2018adversarial} is introduced to iteratively update the perturbations within a limited attack range. 
Subsequently, Madry et al.~\cite{madry2017pgd} utilized local information of the network and employed a min-max optimization formulation(PGD).

Adversarial attacks have been widely applied to various computer vision tasks. For low-level vision tasks, Yin et al.~\cite{yin2018deep} first proposed a generic framework for adversarial attacks for image super-resolution.
Gao et al.~\cite{gao2021advhaze} utilized an atmospheric scattering model to embed potential haze into images as an attack to mislead classifiers into predicting incorrect categories.
Yu et al.~\cite{yu2022towards} conducted a comprehensive evaluation on the robustness of deraining methods and used efficient modules to construct networks against attacks. 
Liu et al.~\cite{liu2023paif} proposed a perceptual fusion framework for multi-modal images and adopted neural network search to balance the robustness of the fusion network and segmentation accuracy.
In the high-level vision domain, Liu et al.~\cite{zhang2019towards} conducted a preliminary exploration on the robustness of the network for object detection, and further developed an adversarial training method for multiple attack sources.
Subsequently, Chen et al.~\cite{chen2021class} adopted class-weighted adversarial training to uniformly enhance the robustness of the detection model. 
Dong et al.~\cite{dong2022adversarially} proposed a robust detector based on adversarial-aware convolutions. 
Wei et al.~\cite{wei2023adversarial} perform attacks through shape and label loss to balance the performance of spectral preservation and attack adversarial, so that the generated pan-sharpened images maintain a high success rate for white-box attacks.
Attacks designed for low-level and high-level vision tasks
have significant differences. Unfortunately, there is no network yet that can defend against both vision-driven and perception-driven attacks simultaneously.

\begin{figure*}[tp]
	\centering
	\setlength{\tabcolsep}{1pt}
	\includegraphics[width=\textwidth]{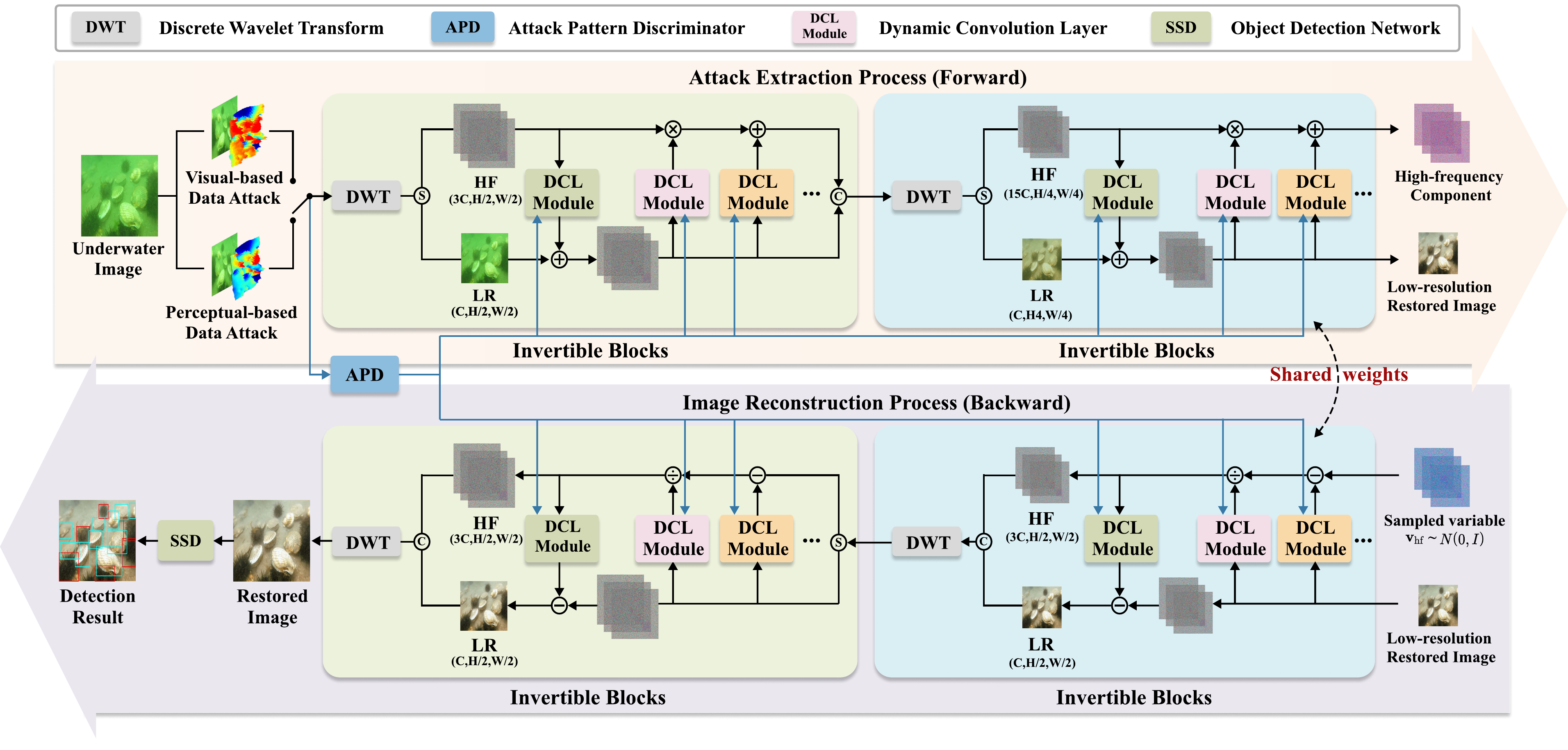}
	\caption{Workflow of the proposed CARNet. In the forward process, the attacked underwater image is transformed into low-resolution image and latent high-frequency component. In the backward process, the high-frequency component is replaced with sampled distribution which is only embedded with clean details to reconstruct enhanced images without attacks.}
	\label{fig:workflow}
\end{figure*}  

\section{Method}
This section starts with a brief introduction to adversarial attacks in underwater tasks. Then, We discuss the implementation process of the introduced bilevel attack optimization strategy. Finally, we provide a clear description of the proposed collaborative adversarsial resilience network.

By incorporating visually imperceptible perturbations $\delta$ into underwater images, the performance of deep learning-based methods can be significantly disrupted. Such perturbations are referred to as underwater adversarial attacks, which can be depicted as follows:
\begin{equation}
	\delta = \underset{\delta}{\arg \max } \;\mathcal{M}\left(\mathbf{r}, f(\mathbf{x} + \delta \mid \omega)\right),
	\label{eq:attack}
\end{equation}
where $\mathbf{x}$ represents the underwater image, $f(\cdot \!\!\mid \!\omega)$ denotes the deep network designed for underwater tasks, which is parameterized by $\omega$. $\mathcal{M}$ means the metric for measuring network performance. $\mathbf{r}$ denotes the task-depend reference information. We hope that by adding perturbation, which maximizes the difference between outputs of the network from the attacked image and the reference. 
To generate adversarial perturbations, widely used optimization-based method PGD~\cite{madry2017pgd} was employed to obtain perturbations iteratively, which can be represented as:
\begin{equation}
	\begin{aligned}
		\delta^{t+1}=\delta^t+\alpha \operatorname{sgn}\left(\nabla_\delta \mathcal{M}\right), \\
	\end{aligned}
	\label{eq:update}
\end{equation}
where $\operatorname{sgn}(\cdot)$ aims to extract the sign of the gradient $\nabla_\delta$. $\alpha$ represents the step size of each iteration. $\delta^0$ is sampled from a uniform distribution and then iteratively updates adversarial attacks on underwater images. Notably, $\delta^{t+1}$ is clipped into $(-\epsilon,\epsilon)$ after each iteration to ensure that the disturbance will not cause obvious visual damage to the input image. $\epsilon$ is a constant value. 
After $t$ iterations, we finally introduce the adversarial attack $\delta^t$ into the input image to evaluate the robustness of the network.

\subsection{Bilevel Attack Optimization Strategy}
Adversarial attacks on underwater images can be divided into two types: vision-driven attack $\delta_v$ and perception-driven attack $\delta_p$. The main difference between them is the design of $M$ from Eq.~\eqref{eq:attack}. Vision-driven adversarial attacks usually employ pixel-level evaluation metrics for underwater image enhancement, such as $l_1$ norm. Perception-driven adversarial attacks usually use metrics for corresponding downstream tasks, such as localization loss and classification loss.

Unlike previous enhancement methods that focus solely on high visual quality, we argue that the image enhancement method has to generate images that are beneficial for both visual inspection and computer perception, known as perception-oriented enhancement. Therefore, either of the attacks mentioned above can compromise certain functionalities of our network.
In order to fortify the proposed network against diverse adversarial attacks, we follow the theory of Stackelberg~\cite{ochs2015bilevel,liu2021investigating,liu2022tardal} and propose a bilevel attack optimization strategy to find the optimal solution, which is as follows:
\begin{equation}
	\begin{gathered}
		\min _{\omega} \mathcal{L}\left(\Psi\left(\omega;\mathbf{u}^*, \delta_p^* \right)\right), \\
		\text { s.t. } \mathbf{u}^* \in \arg \min _{\mathbf{u}}\left( f\left(\mathbf{x}, \delta_v^*\right)\!\rightarrow\! \mathbf{u}\right),\\
		\delta_v^*, \delta_p^* \in \arg \max _{\delta_v, \delta_p}\left(\mathcal{M}_{\mathcal{V}}\left(\delta_v ; \mathbf{x},\mathbf{z} \right), \mathcal{M}_{\mathcal{P}}\left( \delta_p; \mathbf{x},\mathbf{o} \right)\right),
	\end{gathered}
	\label{eq:bilevel}
\end{equation}
where perceptual enhancement and adversarial attack of images represent the upper and lower level respectively. $\mathcal{L}$ denotes the perception-based loss function. $f$ is the vision fidelity term which containing the enhanced image $\mathbf{u}$. 
$\boldsymbol{\omega}$ represents the optimal parameters of the proposed image perception framework $\Psi$. $\mathcal{M}_{\mathcal{V}}$ and  $\mathcal{M}_{\mathcal{P}}$ denote vision-based and perception-based measurement metrics respectively.  
$\mathbf{z}$ and $\mathbf{o}$ represent the clean reference image and the detection label. 
Considering the multiple variables of Eq.~\ref{eq:bilevel}, detection and enhancement are treated as joint problems to mitigate convergence difficulties, which will be described in Section~\ref{APD}.

Notably, we introduce the contrastive principle into the metric $\mathcal{M}_{\mathcal{V}}$ and propose a contrastive attack as vision-driven attacks for underwater image enhancement, which enforces the attacked enhanced results closer to the underwater images and far away from the reference images. The metric for the proposed attack $\mathcal{M}_{\text {CA}}$ is expressed as:
\begin{equation}
	\mathcal{M}_{\text {CA}}= \left\|f(\mathbf{x}+\delta \mid \omega)-\mathbf{z}\right\|_1
	-\left\|f(\mathbf{x}+\delta \mid \omega)-\mathbf{x}\right\|_1,
	\label{eq:enhanceattack}
\end{equation}
where underwater images $\mathbf{x}$ and clean reference images $\mathbf{z}$ denote positive and negative samples respectively. Furthermore, Class Weight Attack~(CWA)~\cite{chen2021class} is introduced as the perception-driven attack for underwater object detection task. The proposed image perception framework $\Psi$ consists of an enhancement module and a detection module, where SSD~\cite{liu2016ssd} is adopted as the backbone of the detection module. The details of the framework will be described in Section~\ref{CarNet}.

\begin{table*}[!htb]
	\renewcommand\arraystretch{2.0}
	\caption{Detailed description of the invertible blocks.}
	\setlength{\tabcolsep}{1.5mm}{
		\begin{tabular}{cccc}
			\hline\textbf{Name} & \textbf{Forward} & \textbf{Backward} & \textbf{Comment} \\
			\hline Step 1 & $\mathbf{u}_1^i, \mathbf{u}_2^i=\operatorname{Split}(\mathbf{u}^i)$ 
			& $\mathbf{u}_1^{i+1}, \mathbf{u}_2^{i+1}=\operatorname{Split}(\mathbf{u}^{i+1})$ 
			& Split feature $\mathbf{u}$ into $\mathbf{u}_1$ and $\mathbf{u}_1$. \\
			\hline \multirow{2}{*}{Step 2}
			& $\mathbf{u}_1^{i+1} = \exp (\mathcal{D}_1(\mathbf{u}_2^{i  }))\odot \mathbf{u}_1^i+\mathcal{D}_2\left(\mathbf{u}_2^i\right)$ 
			& $\mathbf{u}_2^{i  } =(\mathbf{u}_2^{i+1}-\mathcal{D}_4(\mathbf{u}_1^{i+1})) / \exp (\mathcal{D}_3(\mathbf{u}_1^{i+1}))$ 
			& $\mathcal{D}_1$, $\mathcal{D}_2$, $\mathcal{D}_3$ and $\mathcal{D}_4$ are Dynamic Convolution 
			\\
			& $\mathbf{u}_2^{i+1} = \exp (\mathcal{D}_3(\mathbf{u}_1^{i+1}))\odot \mathbf{u}_2^i+\mathcal{D}_4(\mathbf{u}_1^{i+1})$ 
			& $\mathbf{u}_1^i=(\mathbf{u}_1^{i+1}-\mathcal{D}_2(\mathbf{u}_2^i)) / \exp \left(\mathcal{D}_1\left(\mathbf{u}_2^i\right)\right)$ 
			& Layer~(DCL). $\odot$ is the multiply operation. \\
			\hline Step 3 
			& $\mathbf{u}^{i+1}=\operatorname{Concat}(\mathbf{u}_1^{i+1}, \mathbf{u}_2^{i+1})$ 
			& $\mathbf{u}^i=\operatorname{Concat}(\mathbf{u}_1^i, \mathbf{u}_2^i)$ 
			& Merge two features into one in channel-wise. \\
			\hline
	\end{tabular}}
	\label{tab:invertible_process}
\end{table*}

\subsection{Overall Perception Framework}\label{CarNet} 
As for the underwater image enhancement, we denote the attacked underwater image as $\mathbf{y}$ and the clean reference image as $\mathbf{z}$. The degradation factor caused by the physical underwater imaging model~\cite{drews2013formationmodel} is represented as $\varphi$. Therefore, the attack on an underwater image can be expressed as: $p(\mathbf{y}) = p(\mathbf{z}, \varphi,\delta) = p(\mathbf{z})p(\varphi,\delta|\mathbf{z})$. It is obvious that the attacked underwater image contains both potential clean information and degradation information which resulted from physical deterioration and digital attacks. However, it is worth noting that decoupling them and mitigating the interference from both physical and digital sources is a highly complex and challenging endeavor.

The workflow of our method is shown in Fig.~\ref{fig:workflow}. Instead of directly decoupling the clean and degradation information, we aim to separate the low-resolution and high-frequency components of $\mathbf{y}$ by employing wavelet transform and a set of invertible operations.
Considering that the invertible network maintains information integrity~\cite{xiao2020invertiblesr}, we make the first three channels of latent components fit to the clean image after downsampling and fully encode high-frequency information into the remaining channels. Therefore, in the output of the forward process, the features are explicitly decomposed into the low-resolution restored image $\mathbf{x}_{\mathrm{lr}}$ and the high-frequency component $\mathbf{x}_{\mathrm{hf}}$ containing the degradation information, which is expressed as follows: 
\begin{equation}
	p\left(\mathbf{y}\right) =~ p\left(\mathbf{x}_{\mathrm{lr}}, \mathbf{x}_{\mathrm{hf}}, \varphi,\delta\right) = p\left(\mathbf{x}_{\mathrm{lr}}\right) p\left(\mathbf{x}_{\mathrm{hf}}, \varphi,\delta|\mathbf{x}_{\mathrm{lr}}\right).
\end{equation}
During the backward process, we discard all degradation information and reconstruct a clean image from the low-resolution component. However, the high-frequency components obtained by the invertible network contain not only the degradation information but also the clear texture of the clean image. Since it is difficult to effectively decouple them, during the backward process, we replace $\mathbf{x}_{\mathrm{hf}}$ with the sampled $\mathbf{v}_{\mathrm{hf}}\sim N (0, I)$ and combine with $\mathbf{x}_{\mathrm{lr}}$ to reconstruct a clean image $\mathbf{z}$ without degradation information. 
Under the bilateral loss, the clean details of $\mathbf{x}_{\mathrm{hf}}$ are then embedded into the latent variable $\mathbf{v}_{\mathrm{hf}}$ to achieve robust enhancement without suffering from multiple attacks. The bilateral loss on the forward and backward process is illustrated as follows:
\begin{equation}
	\begin{aligned}
		\mathcal{L}_{\text {forw }}& = \frac{1}{M} \sum_{i=1}^M\left\|f_+(\mathbf{y})_{\mathrm{lr}}-\mathbf{z}_{\mathrm{lr}}\right\|_1,\\
		\mathcal{L}_{\text {back }}& = \frac{1}{N} \sum_{i=1}^N\left\|f_-\left(f_+(\mathbf{y})_{\mathrm{lr}}, \mathbf{v}_{\mathrm{hf}}\right)-\mathbf{z}\right\|_1,\\
	\end{aligned}
\end{equation}
where $f_+$ and $f_-$ denote the forward and backward process respectively. $\mathbf{z}_{\mathrm{lr}}$ means the low resolution version of $\mathbf{z}$, which is conducted by bicubic transformation. $M$ and $N$ represent the number of pixels. 

The detailed operations of the invertible blocks are list in Tab.~\ref{tab:invertible_process}. $\operatorname{Split}(\cdot)$ and $\operatorname{Concat}(\cdot)$ denote channel-wise splitting and concatenation respectively. Dynamic convolutional layers~(DCL) are used in this invertible blocks, which is guided by the Attack Pattern Discriminator~(APD) to more effectively denfense different attack types. The detailed description of APD will be illustrated in Section~\ref{APD}.
The SSD detector~\cite{liu2016ssd} is conducted after the clean image is generated to get detection results. The constraint on object detection is as follows:
\begin{equation}
	\mathcal{L}_{\text {det}}=\mathcal{L}_{\text {cls}}+\mathcal{L}_{\text{loc}},
\end{equation}
where $\mathcal{L}_{\text {cls}}$ denotes the classification loss, which is designed to minimize the discrepancy between the predicted and ground truth categories. $\mathcal{L}_{\text {loc}}$ is the localization loss for reducing the difference between the predicted and label boxes.

\begin{figure}[!htb]
	\centering
	\setlength{\tabcolsep}{1pt}
	\includegraphics[width=0.50\textwidth]{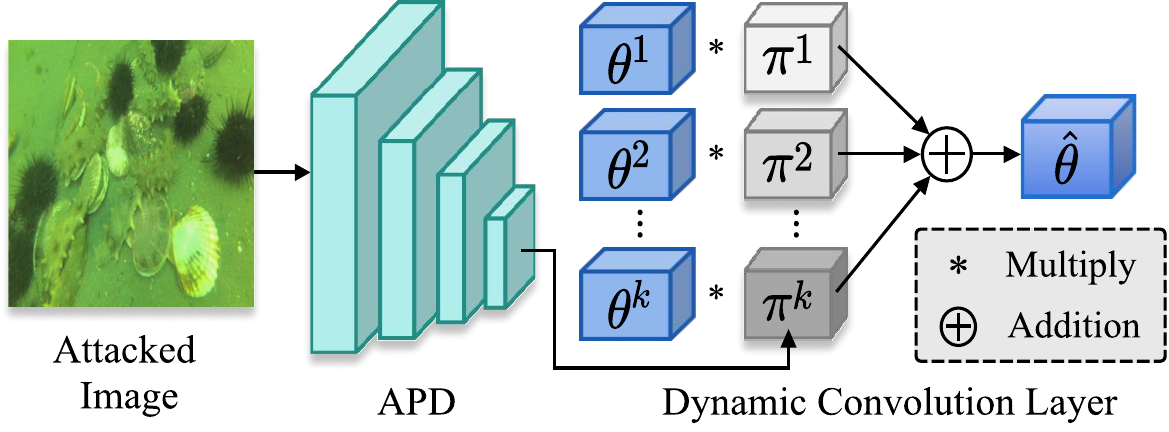}
	\caption{The reweighting process from Attack Pattern Discriminator to Dynamic Convolution Layer. }
	\label{fig:APD}
\end{figure}
\subsection{Attack Pattern Discriminator}\label{APD} 
Different types of attacks show diametrically opposite effects on images. Vision-driven attacks often have limited impact on detection tasks, and vice versa. Therefore, it is difficult for traditional networks with fixed parameters to discern and remove both types of attacks. Inspired by~\cite{chen2020dynamicconv}, instead of employing convolution blocks to learn common features of underwater images under different attacks, we introduce Dynamic Convolution Layer~(DCL) in the invertible network, which aggregates multiple parallel convolution kernels dynamically based upon attention weights. Specifically, the proposed Dynamic Convolution Layer employs different kernels to convolve the images under visual attack and perceptual attack respectively.

Rather than only considering the similarity between visual-driven and perception-driven attacks, the difference between them is additionally learned by proposing an Attack Pattern Discriminator~(APD), which generates input-dependent weights for DCL by identifying the attack type of an input image. The structure of APD is Resnet18~\cite{he2016deep}. The process of APD reweighting DCL is shown in Fig.~\ref{fig:APD}. 
Specifically, the Adversarial Pattern Discriminator~(APD) first generates an $k$-dimensional probability vector $\mathcal{P}=APD(\mathbf{y})=\left\{\pi^1, \pi^2, \ldots, \pi^k\right\}$ based on the type of attack present in the image $\mathbf{y}$. This probability vector serves as a set of weights to modulate the convolution kernels in the dynamic convolution layer. Consequently, the parameters of the resulting convolution kernel $\hat{\theta}_{DCL}$ can be expressed as $\hat{\theta}_{DCL}=\sum_{i=1}^M \pi^i \cdot \theta^i_{DCL} $, where $\theta^i_{DCL}$ denotes the original parameters of the individual dynamic convolution kernels indexed by $i$.

Since APD is also attacked during training, we further employ Online Triplet Constraint~\cite{schroff2015facenet} to APD, which is demonstrated as follows:
\begin{equation}
	\begin{aligned}
		\mathcal{L}_{\mathrm{APD}}&=\sum_{i=1}^{N_T}(J S\left(APD\left(\mathbf{x}_i^a\right), APD\left(\mathbf{x}_i^p\right)\right)-\\
		&J S\left(APD\left(\mathbf{x}_i^a\right), APD\left(\mathbf{x}_i^n\right)\right)+\gamma)_{+},
	\end{aligned}
\end{equation}
where $\mathbf{x}_i^p$ and $\mathbf{x}_i^n$ represents the corresponding positive and negative pairs of $\mathbf{x}_i^a$, $\gamma$ is the margin that is enforced between $\mathbf{x}_i^a$ and $\mathbf{x}_i^p$. $M$ denotes the number of triplets. $JS$ represents Jensen-Shannon divergence~\cite{lin1991divergence} of two distributions, which can be illustrated as $J S\left(P_1, P_2\right)=\frac{1}{2} K L\left(P_1 \| \frac{P_1+P_2}{2}\right)+\frac{1}{2} K L\left(P_2 \| \frac{P_1+P_2}{2}\right)$,
where $K L$ represents Kullback-Leibler Divergence~\cite{van2014renyi}. $P_1$ and $P_2$ denotes the corresponding probability distributions.
Therefore, the whole loss during the training process can be expressed as:
\begin{equation}
	\mathcal{L}_{\mathrm{train}}=\lambda_1\mathcal{L}_{\text {forw }}+\lambda_2\mathcal{L}_{\text {back }}+\lambda_3\mathcal{L}_{\text {det }}+\lambda_4\mathcal{L}_{\mathrm{APD}},
\end{equation}
where $\lambda_1$, $\lambda_2$, $\lambda_3$ and $\lambda_4$ represents the hyper-parameters which controls the balcance of above terms respectively.
\begin{figure*}[!htb]
	\centering
	\setlength{\tabcolsep}{1pt}
	\includegraphics[width=0.99\textwidth]{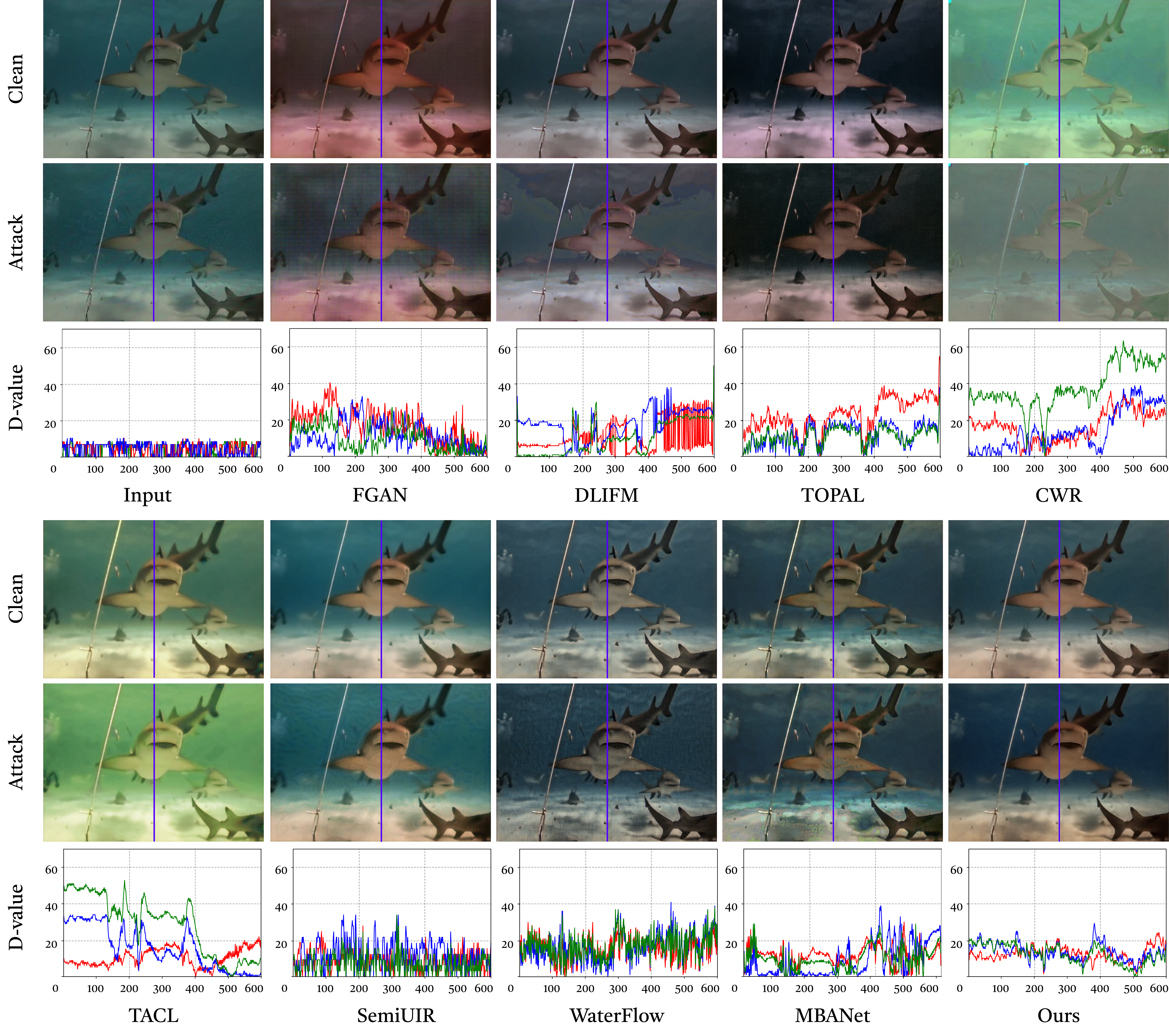}
	\caption{Enhancement results on UIEBD dataset. The line graph represents the image difference between the clean image and the attack image after enhancement. Smaller differences indicate better robustness.}
	\label{fig:visualizationforuiebd}
\end{figure*}
\begin{figure*}[!htb]
	\centering
	\setlength{\tabcolsep}{1pt}
	\includegraphics[width=0.99\textwidth]{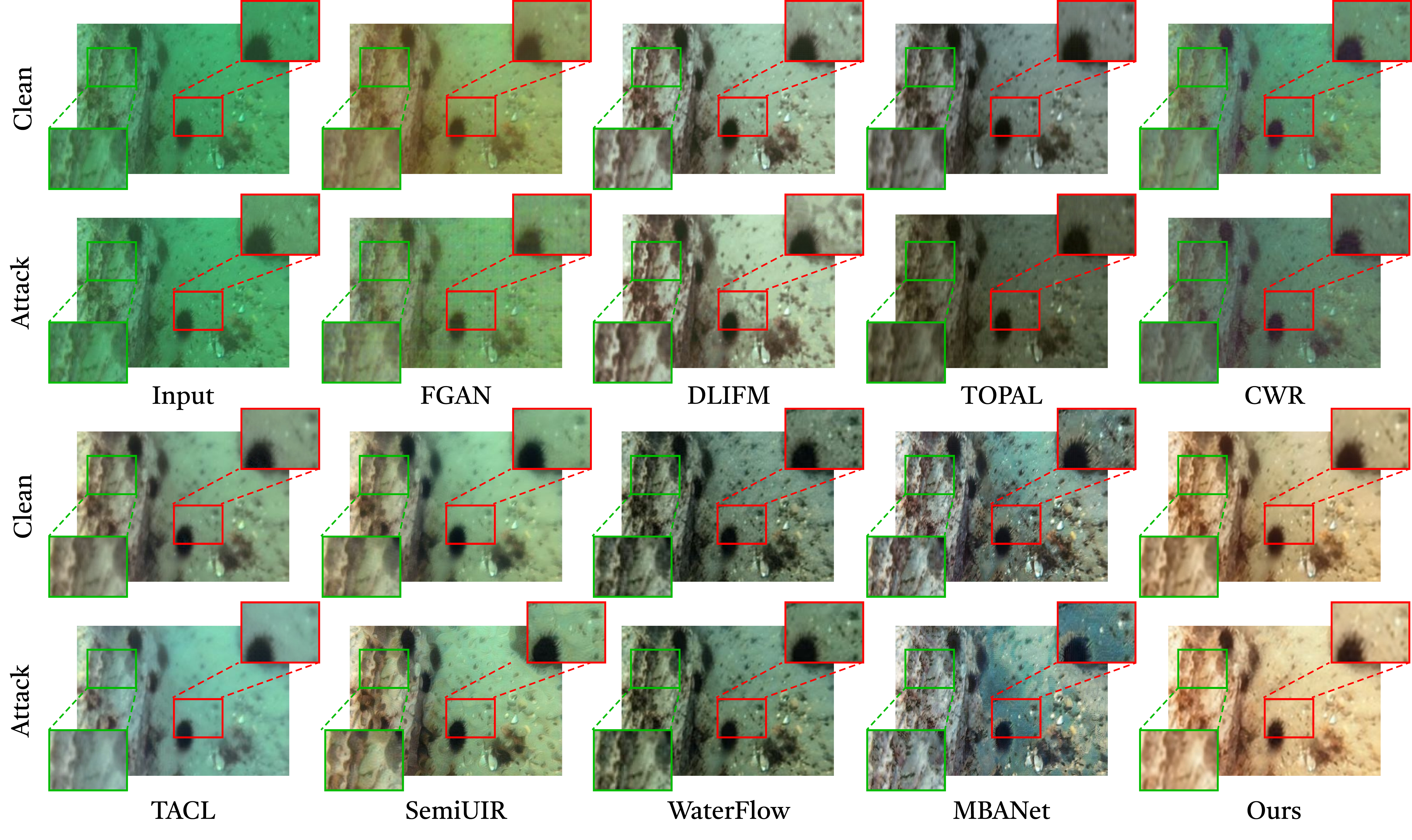}
	\caption{Enhancement results on RUIE dataset.}
	\label{fig:visualizationforuccs}
\end{figure*}
\begin{table*}[!htb]
	\renewcommand\arraystretch{1.35}
	\caption{Quantitative comparison for underwater image enhancement without adversarial attacks in terms of UCIQE($\uparrow$), UISM($\uparrow$), CEIQ($\uparrow$), HIQA($\uparrow$), PSNR($\uparrow$) and SSIM($\uparrow$). $\uparrow$ denotes that large values means better results. The best and second results are marked in \textbf{bold} and \underline{underline}.}
	\setlength{\tabcolsep}{2.9mm}{
		\begin{tabular}{l|cccccc|cccc}
			\toprule 
			\multirow{2}{*}{\textbf{Method}}
			&\multicolumn{6}{c|}{\textbf{UIEBD}~\cite{li2019waternetuiebd}}
			&\multicolumn{4}{c}{\textbf{RUIE}~\cite{liu2020uccs}}\\
			&UCIQE($\uparrow$)&UISM($\uparrow$)&CEIQ($\uparrow$)&HIQA($\uparrow$)&PSNR($\uparrow$)&SSIM($\uparrow$)
			&UCIQE($\uparrow$)&UISM($\uparrow$)&CEIQ($\uparrow$)&HIQA($\uparrow$)\\
			\midrule 
			FGAN 
			&  0.5457          &  2.7378            &  3.0825       &0.2954  
			&  15.7601         &  0.6731       
			&  \underline{0.5235}          & 2.5008           &  2.8275      
			&  0.3071 \\
			DLIFM  
			& 0.5742       &2.7805           &  3.2424    &0.3069  
			&  17.2531         & 0.7824           
			&  0.5103          & 3.5286            &  2.9511        
			&  0.3228                       
			\\TOPAL  
			&  0.5644            & 2.7240     & 3.1999          &0.2677 
			&  16.9008            &  0.6939 
			&  0.4740            &  2.8977            & 3.0402   
			&  0.2903           
			\\CWR
			&  0.4782          & 2.4707      & 2.7286         &\textbf{0.3164} 
			&  15.8725            &  0.7430  
			&  0.4922          &  3.2954      & 2.3257   
			&  0.3133           
			\\TACL  
			&  0.5474          & 2.3804            & 3.2921   &0.2841 
			&  17.9835          & 0.7637 
			&  0.5026        & 2.8015            & 2.9063
			&  0.2826            
			\\SemiUIR  
			&  \underline{0.5747}    & 2.7267            & 3.2784   &0.3052 
			&  \textbf{19.1920}          & \textbf{0.8328} 
			&  \textbf{0.5355}        & 2.8916            & 2.8944
			&  0.2933            
			\\WaterFlow  
			&  0.5725          &  \underline{2.9557}         & 3.3027  &0.2916 
			&  18.5186          & 0.8174 
			&  0.5214        & \underline{3.5834}            & 2.9661
			&  0.3228            
			\\MBANet  
			&  0.5687          &2.8055           & \underline{3.3096}   &0.2909 
			&  19.0419          & 0.8053 
			&  0.5026        & 3.1116            & \textbf{3.4634}
			&  \underline{0.3974}            
			%
			\\Ours     
			& \textbf{0.5981}   & \textbf{3.0516}   & \textbf{3.3408} & \underline{0.3082} 
			& \underline{19.1315}   &    \underline{0.8228}
			
			& 0.5190   & \textbf{3.7099}   & \underline{3.0986}
			& \textbf{0.4087}        
			\\ 
			\bottomrule
	\end{tabular}}
	\label{tab:evaluation_enhance_unattack}
\end{table*}

\begin{table*}[!htb]
	\renewcommand\arraystretch{1.35}
	\caption{Quantitative comparison for underwater image enhancement under adversarial attacks.}
	\setlength{\tabcolsep}{2.9mm}{
		\begin{tabular}{l|cccccc|cccc}
			\toprule 
			\multirow{2}{*}{\textbf{Method}}
			&\multicolumn{6}{c|}{\textbf{UIEBD}~\cite{li2019waternetuiebd}}
			&\multicolumn{4}{c}{\textbf{RUIE}~\cite{liu2020uccs}}\\
			&UCIQE($\uparrow$)&UISM($\uparrow$)&CEIQ($\uparrow$)&HIQA($\uparrow$)&PSNR($\uparrow$)&SSIM($\uparrow$)
			&UCIQE($\uparrow$)&UISM($\uparrow$)&CEIQ($\uparrow$)&HIQA($\uparrow$)\\
			\midrule 
			FGAN 
			&  0.5390            &  2.9233            &  3.0284          &  0.2582   
			&  15.3797            &  0.6305        
			&  0.4851            &  3.2198            &  2.7385    
			&   0.2449                  
			\\ DLIFM  
			&  0.5693           &  2.8001           &  3.2034   &  0.3024      
			&  16.6299 &  0.7111         
			&  0.5074             &  3.7745             &  2.9318        
			&  0.3258                      
			\\ TOPAL  
			&  0.5285            & \underline{3.0628}   & 3.0608         &  0.2888 
			&  15.4115           & 0.5758 
			&  0.4659            & 3.1195            & 3.0271  
			&  0.2915            
			\\ CWR
			&  0.4674          & 2.0309       & 2.3632   &  \textbf{0.3100}
			&  13.1063         &  0.5902        
			&  0.4565          &  2.7451      & 2.2299   
			&  0.3252           
			\\ TACL  
			&  0.5251  & 2.2383  & 3.2037  &  0.2810 
			& 15.7862 & 0.7031
			&  0.5106         & 2.6630       & 2.9686
			&  0.2761      
			\\ SemiUIR  
			&  \underline{0.5774}  & 3.8524  & 3.2655  &  0.3045 
			& \underline{17.6756} & 0.7703
			&  \underline{0.5125}         & 3.4159       & 2.8793 
			&  0.3590        
			\\ WaterFlow  
			&  0.4989  & 3.0276  & 3.2546  &  0.2958 
			& 16.5697 & 0.7070
			&  0.4926         & \underline{3.8768}       & 2.9753
			&  0.3077        
			\\ MBANet  
			&  0.5035  & 2.9834  & \underline{3.3065}  &  0.2905 
			& 17.1832 & \underline{0.7874}
			&  0.5095        & 3.1218       & \textbf{3.4562}
			&  \underline{0.4182}            
			\\Ours     
			& \textbf{0.5829}   & \textbf{3.0662}   & \textbf{3.3207}&  \underline{0.3068}
			& \textbf{17.8854}   &  \textbf{0.7927}  
			& \textbf{0.5131}   & \textbf{3.9034}   & \underline{3.0301}
			&  \textbf{0.4231}           
			\\ 
			\bottomrule
	\end{tabular}}
	\label{tab:evaluation_enhance_attack}
\end{table*}

\begin{table*}[!htb]
	\renewcommand\arraystretch{1.3}
	\caption{Quantitative comparison for underwater object detection in terms of mAP($\uparrow$).}
	\setlength{\tabcolsep}{1.56mm}{
		\begin{tabular}{l|ccccc|ccccc|ccccc}
			\toprule 
			\multirow{2}{*}{\textbf{Method}}
			&\multicolumn{5}{c|}{\textbf{RUIE}~\cite{liu2020uccs}}
			&\multicolumn{5}{c|}{\textbf{Aquarium}~\cite{aquarium}}
			&\multicolumn{5}{c}{\textbf{UTDAC2020}~\cite{chen2022utdac}} \\
			&$\mathbf{Clean}$&$\mathbf{A_{cls}}$&$\mathbf{A_{loc}}$&$\mathbf{CWT}$&$\mathbf{DAG}$
			&$\mathbf{Clean}$&$\mathbf{A_{cls}}$&$\mathbf{A_{loc}}$&$\mathbf{CWT}$&$\mathbf{DAG}$
			&$\mathbf{Clean}$&$\mathbf{A_{cls}}$&$\mathbf{A_{loc}}$&$\mathbf{CWT}$&$\mathbf{DAG}$
			\\
			\midrule 
			SSD    
			&\textbf{91.36}& 10.70      & 13.46     & 9.90     & 42.63
			&\textbf{55.42}& 2.61       & 2.84      & 0.97     & 15.76
			&\textbf{67.30}& 4.03       & 5.91      & 0.22     & 7.94\\
			SSD-AT    
			& 57.42        & 38.74      & 44.18     &40.28     & 46.67
			& 26.80        & 16.36      & 24.04     & 19.92    & 21.20
			& 16.15        &  9.74      & 13.98     & 14.33    & 11.85\\
			DLIFM  
			& 65.86        & 43.62      & 44.31     & 49.63    & 52.16
			& 43.08        & 16.19      & 17.12     &15.20     & 32.24
			& 22.34        & 13.22      & 14.52     & 16.67    & 18.26
			\\FGAN 
			& 76.58        & 38.01      & 38.92     & 42.27    & 46.51
			& 46.26        & 17.52      & 24.00     & 18.21    & 23.25
			& 26.17        & 13.51      & 13.98     & 16.86    & 19.90
			\\TOPAL  
			& 79.03        & 72.52      & 67.96     & 62.40    & 69.16
			& 47.54        & 24.53      & 27.41     & 25.02    & 31.86
			& 26.34        & 14.96      & 15.13     & 17.92    & 20.32
			\\CWR
			& 80.64        & 72.20      & 72.92     & 68.62    & 74.52
			& 46.51        & 21.17      & 23.68     & 22.77    & 29.81
			& 28.52        & 14.82      & 15.43     & 16.56    & 18.66
			\\TACL  
			& 82.21          & \underline{76.69}    & \underline{78.14}   & 72.66& \underline{79.26}
			& 47.36          & \underline{26.10}         & \underline{29.44}        & \underline{27.12}  & \underline{36.28}
			& 31.17           & \underline{16.28}          & \underline{16.92}       & \underline{18.00}    & \underline{21.26}
			\\SemiUIR  
			& 81.57          & 74.57    & 77.41   & \underline{72.75}  & 79.11
			& 47.21          & 24.93    & 27.43   & 26.57  & 36.28
			& 31.17          & 16.28    & 16.92   & 18.00  & 21.26
			\\WaterFlow
			& 82.18          & 73.98    & 75.39   & 70.57  & 77.04
			& 46.71          & 24.48    & 26.92   & 25.88  & 35.82
			& 31.17          & 16.28    & 16.92   & 18.00  & 21.26
			\\MBANet
			& 79.33          & 71.65    & 73.15   & 69.89  & 77.46
			& 46.03          & 23.59    & 25.41   & 25.32  & 35.01
			& 31.17          & 16.28    & 16.92   & 18.00  & 21.26
			\\Ours     
			&\underline{90.30}     & \textbf{84.61}      & \textbf{84.83}         & \textbf{81.14}          & \textbf{89.85}
			&\underline{51.79}           & \textbf{32.63}          & \textbf{39.15 }        & \textbf{32.01 }         & \textbf{45.26}
			& \underline{44.32}          & \textbf{18.86}          & \textbf{18.37}         & \textbf{21.58 }          & \textbf{24.97}
			\\ 
			\bottomrule
	\end{tabular}}
	\label{tab:evaluation_detection}
\end{table*}
\section{Experiments}
\subsection{Datasets and Metrics}
In this section, we evaluate the effectiveness of the poprposed method on underwater enhancement and subsequent detection tasks through quantitative and qualitative experiments. Specifically, widely used underwater datasets UIEBD~\cite{li2019waternetuiebd} and RUIE~\cite{liu2020uccs} are used to assess the performance of our method on underwater enhancement tasks. RUIE~\cite{liu2020uccs}, Aquarium~\cite{aquarium}, and UTDAC2020~\cite{chen2022utdac} datasets are used for evaluating our method on underwater object detection tasks.

The UIEBD dataset~\cite{li2019waternetuiebd} consists of 890 image pairs, of which 800 are allocated for training and 90 for testing. 
The dataset is derived partly from online repositories and partly through proactive collection underwater. 
To facilitate more precise evaluations, the underwater images have been processed with 12 different enhancement algorithms, allowing volunteers to select the optimal images to be used as reference standards. 
Meanwhile, the RUIE dataset~\cite{liu2020uccs} includes 2,074 training pairs and 675 test pairs, complete with detection labels. This dataset is typically utilized to assess the performance of underwater image enhancement and object detection tasks.
The detection labels encompass three representative types of seafood: urchins, trepangs, and scallops. 
The aquarium dataset~\cite{aquarium} was collected from two aquariums in the United States and includes a total of 511 images, consisting of 448 training images and 63 test images. 
The dataset is labeled with seven common underwater animals: fish, jellyfish, penguins, sharks, puffins, stingrays, and starfish. 
The UTDAC2020 dataset~\cite{chen2022utdac} is proposed for the Underwater Object Detection Algorithm Competition, which comprises 5,168 training images and 1,293 validation images, covering four categories: echinus, holothurian, starfish, and scallop. 
This dataset is available in four resolutions: 3840$\times$2160, 1920$\times$1080, 720$\times$405, and 586$\times$480.

Underwater Color Image Quality Evaluation~(UCIQE)~\cite{yang2015uciqe}, Underwater
Image Sharpness Measure~(UISM)~\cite{panetta2015uiqmuism}, Contrast
Enhancement based Contrast-changed Image Quality Measure~(CEIQ)~\cite{yan2019ceiq} and Self-adaptive Hyper Network Architecture~(HIQE)~\cite{su2020hiqa} are employed as unsupervised image quality evaluation metrics. UCIQE evaluates color cast, blurriness, and contrast deficiencies in underwater images, which provides an efficient and swift solution for real-time image processing, and improving alignment with subjective visual perception evaluations. 
UISM evaluates the sharpness of underwater images, indicating the effectiveness of detail and edge preservation duiring the enhancement process.
CEIQ is designed to measure the quality of contrast-distorted images by evaluating the similarity between the input image and its enhanced version.
HIQA introduces a hypernetwork to adaptively adjust quality prediction parameters to obtain a perceptual effect that is closer to the way humans understand the world.
Additionally, Peak Signal to Noise Ratio~(PSNR) and Structural Similarity~(SSIM)~\cite{wang2004ssim} are used as full-reference metrics on the UIEBD dataset with reference images. Higher values of these metrics indicate better image quality.

The mean Average Precision (mAP) is adopted as the evaluation metric to measure the detection accuracy of the current method. Widely used object detection adversarial attacks, including traditional attacks $A_{cls}$, $A_{loc}$, and well-crafted attacks CWA~\cite{chen2021class}, DAG~\cite{xie2017adversarial} are employed to comprehensively evaluate the robustness of our method. 
For fair comparisons, we select recent representative learning-based enhancement methods FGAN~\cite{islam2020fgan}, DLIFM~\cite{chen2021dlifm}, TOPAL\cite{jiang2022topal}, CWR~\cite{han2022cwr}, TACL~\cite{liu2022tacl},  SemiUIR~\cite{huang2023contrastive}, WaterFlow~\cite{zhang2023waterflow} and MBANet~\cite{xue2023investigating} as our comparative methods. Additionally, we subject each method to the same adversarial training strategy with SSD detector to demonstrate the robustness of the proposed network in an intuitive manner.

\subsection{Implement Details}
The proposed network is implemented in Pytorch and trained on NVIDIA RTX 4090 GPU. We first pre-train the enhancement part for 300,000 iterations and the detection part for 120000 iterations separately. UIEBD dataset is used as the training dataset for underwater image enhancement. RUIE, Aquarium and UTDAC2020 are used as the training dataset for object detection, respectively. Subsequently, we jointly train for 200,000 iterations on image enhancement and object detection tasks by using the proposed training strategy. During the training process, SGD is employed as the optimizer for our network, with the learning rate set to 1e-4. $\lambda_1$, $\lambda_2$, $\lambda_3$, $\lambda_4$ is set to 2, 0.1, 1, 5 respectively.

\begin{figure*}[!htb]
	\centering
	\setlength{\tabcolsep}{1pt}
	\includegraphics[width=0.99\textwidth,height=0.23\textheight]{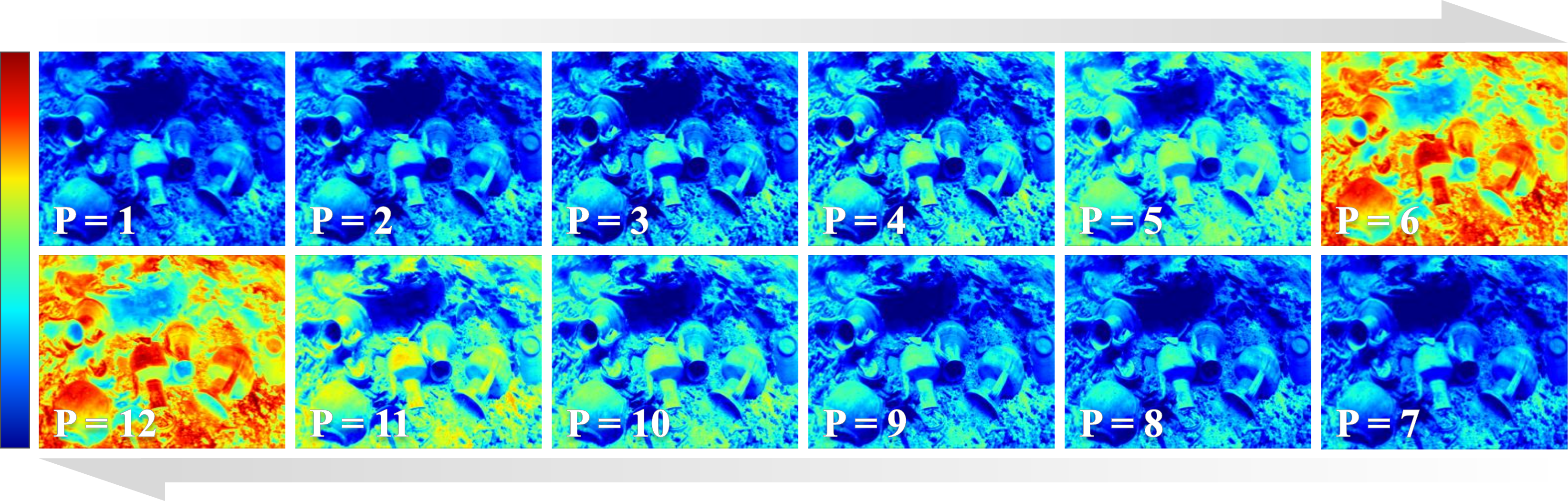}
	\caption{Heatmap for study of the invertible process.}
	\label{fig:heatmap}
\end{figure*}
\begin{figure*}[!htb]
	\centering
	\setlength{\tabcolsep}{1pt}
	\includegraphics[width=0.99\textwidth]{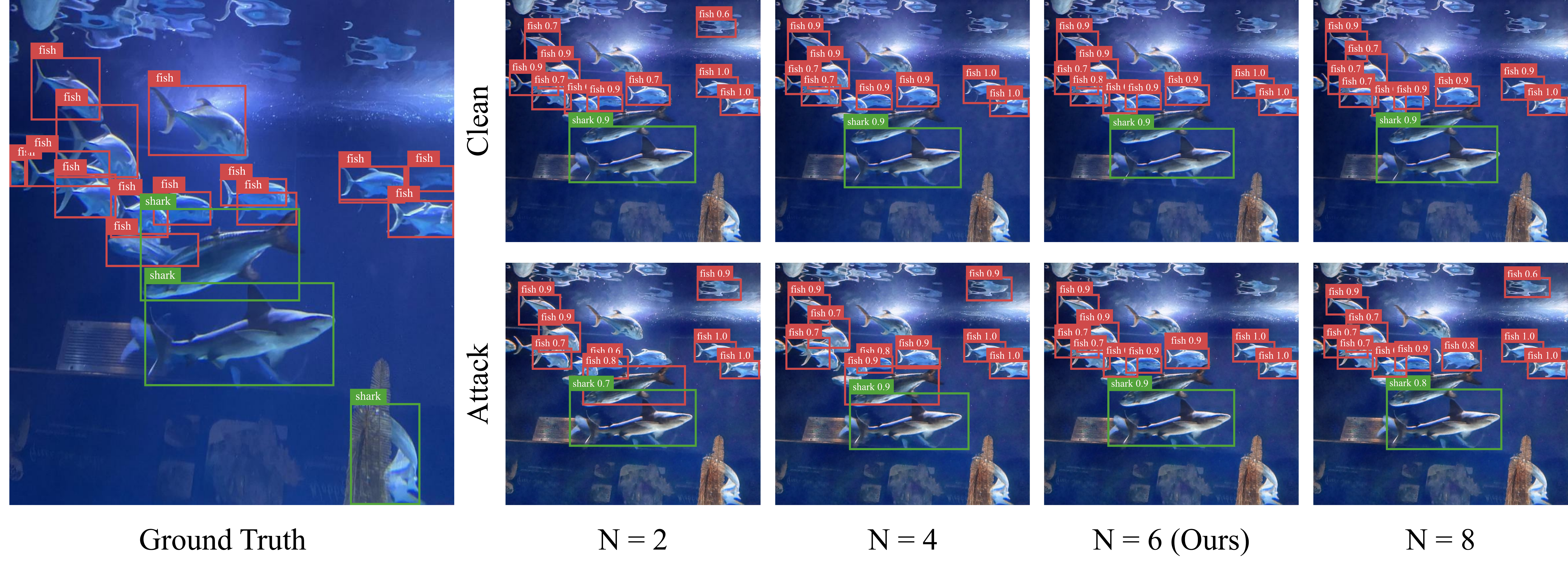}
	\caption{Detection results of ablation study on the number of Invertible Blocks.}
	\label{fig:visualizationforaquarium}
\end{figure*}
\begin{figure*}[!htb]
	\centering
	\setlength{\tabcolsep}{1pt}
	\includegraphics[width=0.99\textwidth,height=0.24\textheight]{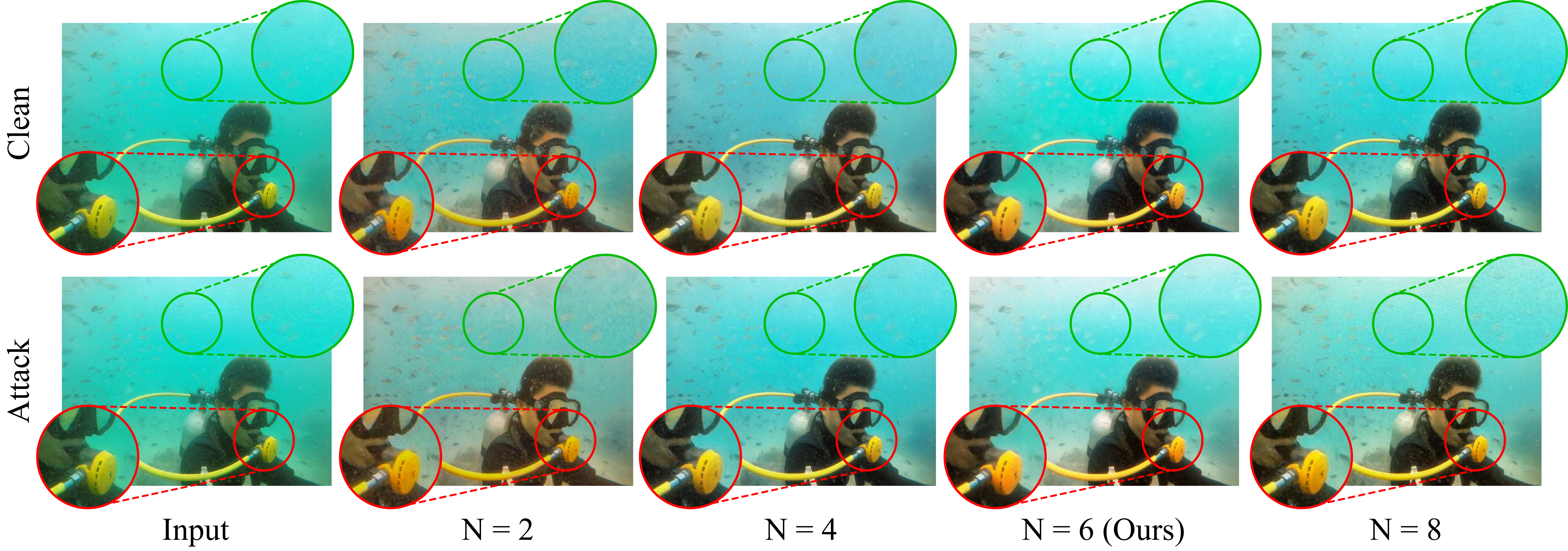}
	\caption{Enhancement results of ablation study on the number of Invertible Blocks.}
	\label{fig:visablationblocks}
\end{figure*}
\begin{figure*}[!htb]
	\centering
	\setlength{\tabcolsep}{1pt}
	\includegraphics[width=0.99\textwidth,height=0.25\textheight]{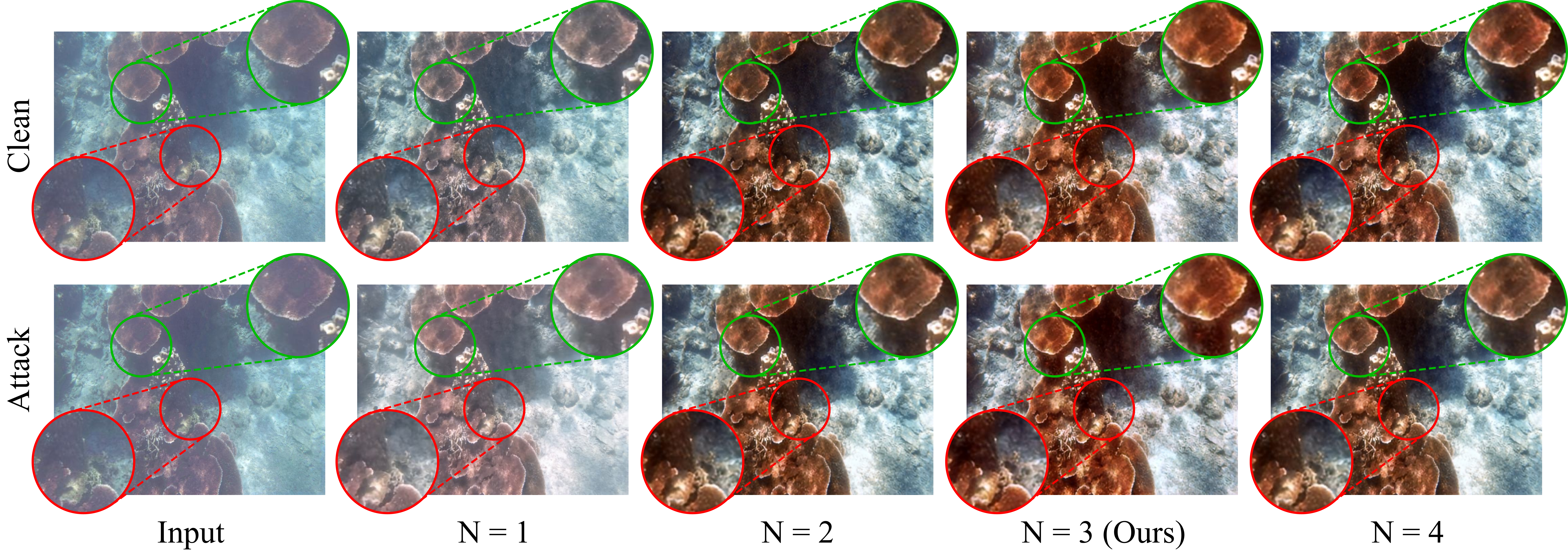}
	\caption{Enhancement results of ablation study on the number of Dynamic Convolutional Layers.}
	\label{fig:visablationdcls}
\end{figure*}
\begin{figure*}[!htb]
	\centering
	\setlength{\tabcolsep}{1pt}
	\includegraphics[width=0.99\textwidth]{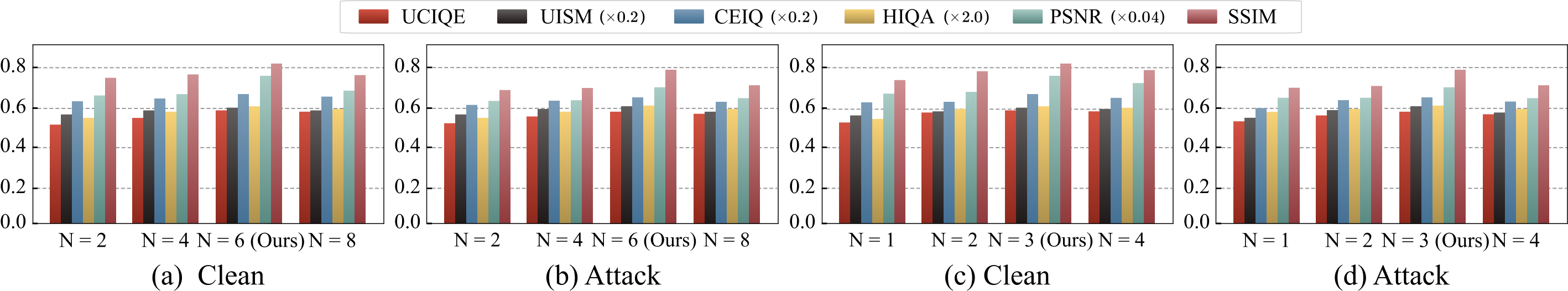}
	\caption{Quantitative results of ablation study on visual quality enhancement tasks. (a) and (b) illustrate the ablation study on the number of Invertible Blocks under clean and attacked scenarios, respectively. (c) and (d) present the ablation study on the number of Dynamic Convolutional Layers in both clean and attacked scenarios.}
	\label{fig:ablationzhuxinghtu}
\end{figure*}
\subsection{Qualitative Results}
We first evaluate the performance of our method on the underwater image enhancement task. Fig.~\ref{fig:visualizationforuiebd} shows the performance of our method on the UIEBD dataset. The results in the first and second rows represent the enhancement results obtained by all enhancement methods without and with adversarial attacks, respectively. 
It is apparent that in the scene of the image example, the red and green light have severely degraded, leading to a noticeable blue color cast in the final image. From the results without attacks, CWR and TACL show limited enhancement effects on underwater images. FGAN produces additional artifacts which make the result deviate from the original color. While TOPAL restores a good color appearance to the image, it suffers from dark reflection. WaterFlow fails to generate bright reflections. Only SemiUIR, MBANet, and the proposed method effectively restore the original colors of the underwater images.

After introducing adversarial attacks, the enhancement effects of CWR and TACL are further weakened after being attacked. DLIFM is affected by local over-enhancement, which destroys the overall structural appearance of the image. FGAN is disturbed by obvious noise, resulting in artifacts and unnatural colors. The previously observed issue of excessive dimness in TOPAL is further exacerbated after being attacked. SemiUIR, WaterFlow, and MBANet, while enhancing image contrast, also accentuate the interference from minor perturbations. In contrast, only the proposed method generates relatively satisfactory images both unattacked and attacked. We further evaluate whether the enhanced results from the unattacked and the attacked inputs have appearance consistency. 
The line chart in the third row represents the absolute value of the intensity difference between the unattacked and the attacked enhanced image on the RGB channels. It is evident that all enhancement methods have been affected by the perturbations. 
Compared with other enhancement methods, only the proposed method suffers the least perturbation, which proves the robustness of our method to attacks.

Fig.~\ref{fig:visualizationforuccs} shows the enhancement effect of the proposed method on the RUIE dataset. Unlike the example from Fig.~\ref{fig:visualizationforuiebd}, this scene suffers from significant green artifacts. While FGAN, DLIFM and CWR show some improvement in image contrast, they lack effectiveness in fully eliminating color casts and artifacts. After being attacked, they show more obvious color artifacts. 
TOPAL and WaterFlow suffer from low saturation, which is exacerbated after introducing attacks. Adversarial training makes TACL lose the details of the content during the enhancement process, resulting in blurred enhanced images. 
In contrast, our images not only generate bright reflections but also recover richer detail information.

\begin{figure*}[!htb]
	\centering
	\setlength{\tabcolsep}{1pt}
	\includegraphics[width=0.99\textwidth]{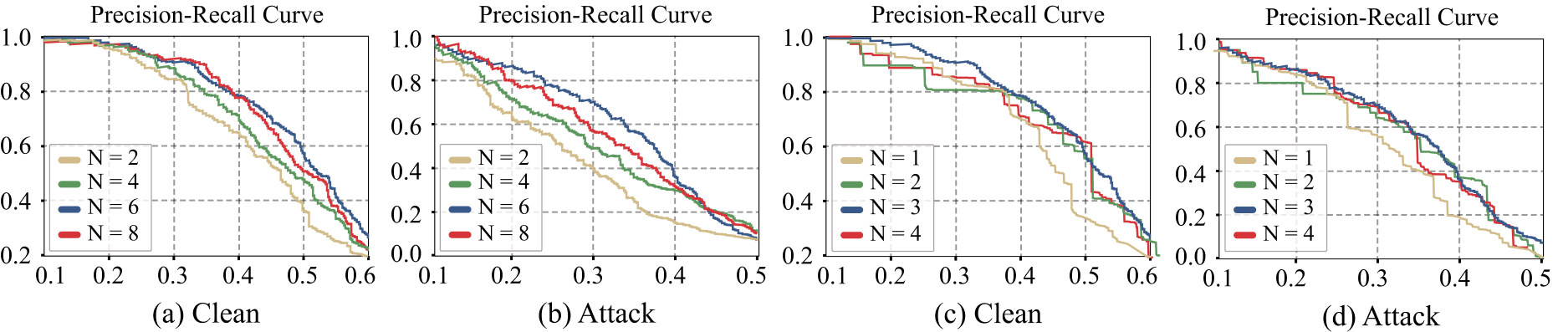}
	\caption{Quantitative results of ablation study on detection tasks.~(a) and (b) illustrate the ablation study on the number of Invertible Blocks under clean and attacked scenarios, respectively.~(c) and (d) present the ablation study on the number of Dynamic Convolutional Layers in both clean and attacked scenarios.}
	\label{fig:ablationpr}
\end{figure*}
\subsection{Quantitative Results}
Table~\ref{tab:evaluation_enhance_unattack} presents the quantitative results of the proposed method without attacks. On the UIEBD dataset, our method outperforms existing methods in all metrics except HIQA, PSNR and SSIM. PSNR and SSIM are slightly behind SemiUIR. HIQA is closely behind CWR. On the RUIE dataset, our method achieved the best performance in UISM and HIQA, and is lower than SemiUIR and WaterFlow in UCIQE and CEIQ. Table~\ref{tab:evaluation_enhance_attack} shows the quantitative results under attacks. It can be clearly seen that the advantage of our method is further enhanced. On the UIEBD dataset, compared with the clean version, it surpasses all comparison methods in both PSNR and SSIM and the gaps with the highest metrics HIQA decreased from 2.67\% to 1.04\%. On the RUIE dataset, only CEIQ remained lower than MBANet, while all other metrics surpassed other methods.

Table~\ref{tab:evaluation_detection} demonstrates the quantitative results of the proposed method on three detection datasets. SSD represents the detection results without using any enhancement network for joint training and without adversarial training. SSD-AT is the adversarial-trained version of SSD. It can be observed that when no adversarial training is conducted, the mean average precision (mAP) on clean images is higher than all existing methods. However, the mAP drops significantly once any attack is applied. Compared with existing methods, the proposed method shows the least difference with pure SSD in test performance on clean images. Meanwhile, the precision on the attacked underwawer images is significantly higher than all other methods.

\begin{figure}[!htb]
	\centering
	\setlength{\tabcolsep}{1pt}
	\includegraphics[width=0.47\textwidth]{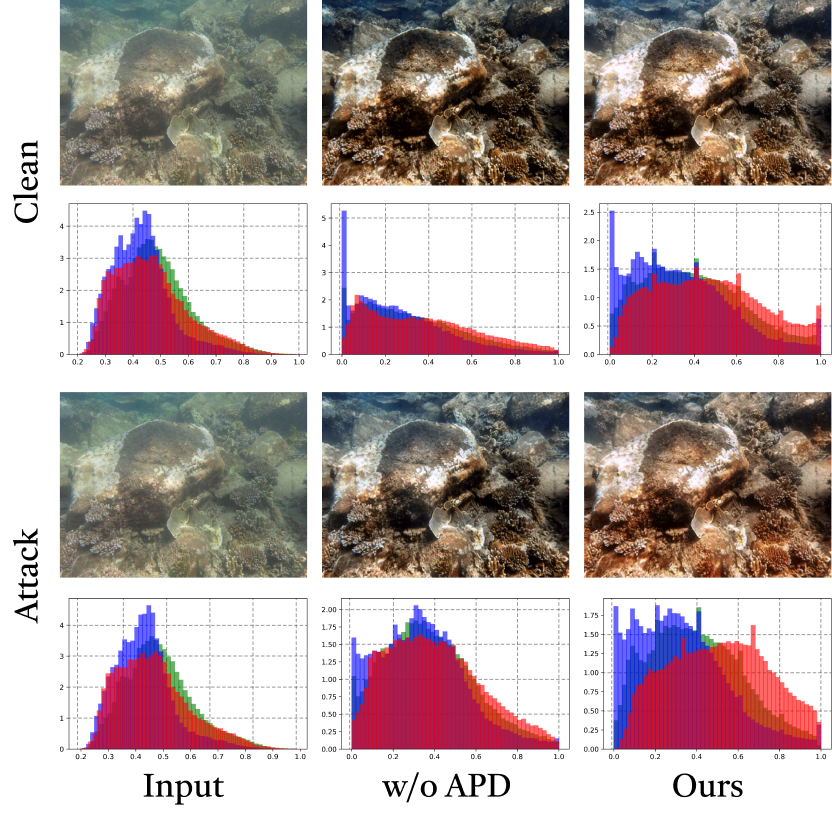}
	\caption{Enhancement results of ablation study on the Attack Pattern Discriminator.}
	\label{fig:visualizationforapd}
\end{figure}

\subsection{Ablation Study}
\subsubsection{Study on Number of Invertible Blocks}
The proposed framework consists of multiple invertible blocks. We further explored the influence of the number of Invertible Blocks for our method. The visualization of enhancement results is shown in Fig.~\ref{fig:visablationblocks}. It is evident that the enhancement results when N = 2 shows a certain extent of color casting. When N = 8, the introduction of the attack introduces additional noise interference in the enhanced image, especially noticeable in the content within the green box. Compared with N = 4, which produces a relatively vivid appearance, N = 6 substantially radiance and contrast of the scene. 

The quantitative results on visual enhancement are shown in (a) and (b) of Fig.~\ref{fig:ablationzhuxinghtu}, where (a) and (b) represent the enhancement results without and with attack. When testing on clean images, N = 6 reach their optimum in all metrics. After being attacked, N = 6 still outperforms all other configurations in all metrics. (a) and (b) in Fig.~\ref{fig:ablationpr} denote the detection comparison on clean and attacked images, in which the detection result when N = 6 is better than other results on clean images. Simultaneously, when N = 6, the interference is significantly less than in other cases under attacked underwater images. Therefore, we ultimately set N = 6 in our method.
\subsubsection{Study on Number of Dynamic Convolutional Layers}
Each of the proposed Invertible Blocks consists of multiple Dynamic Convolutional Layers~(DCLs). We study the number of dynamic convolutional layers to find the optimal solutions for robust image perception. The enhanced results of our method are shown in (c) and (d) in Fig.~\ref{fig:visablationdcls}. We use N to represent the number of DCLs. It can be seen that in this example, the attack has limited impact on the visual effect of the image. It is obvious that when N = 1, the scant network parameters resulting in limited recovery effect. When N is greater than 2, the proposed methods effectively estimate the reflection of the underwater scene, resulting in good enhancement effect. However, compared with N = 2 or 4, when N = 3, the foreground objects in the image are better restored, especially in the green frame scene where the coral colony shows a more vivid color effect. 

The visual enhancement effects on clean and attacked images are shown in (c) and (d) of Fig.~\ref{fig:ablationzhuxinghtu}. When N = 3, the proposed method achieves a significant advantage in both clean and attacked images. (c) and (d) in Fig.~\ref{fig:ablationpr} show the quantitative results of object detection tasks on clean and attacked images. When testing clean images, the detection results when N = 3 are significantly higher than those when N = 1, 2, or 4. When testing attack images, although the detection effect is lower than that of clean images, it still shows outstanding advantages compared with other cases. Therefore, we finally refer to 3 dynamic convolutional layers in the reversible block.
\begin{figure}[!htb]
	\centering
	\setlength{\tabcolsep}{1pt}
	\includegraphics[width=0.48\textwidth,height=0.13\textheight]{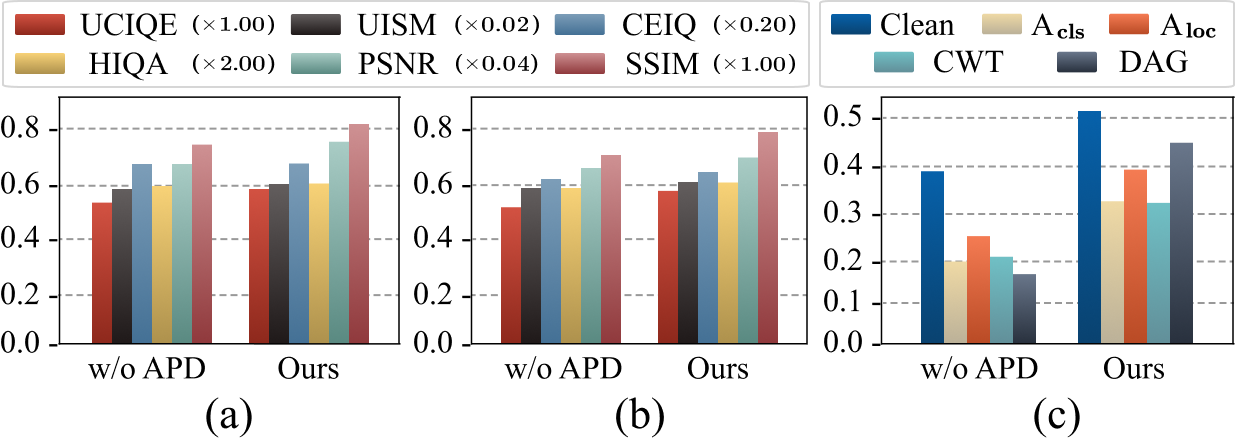}
	\caption{Quantitative results of ablation study on the Attack Pattern Discriminator. (a) and (b) demonstrate the study on enhancement effect, and (c) shows the study on object detection performance.}
	\label{fig:ablationapd}
\end{figure}
\subsubsection{Study on Attack Pattern Discriminator}
To verify the performance of the Attack Pattern Discriminator~(APD), we investigated the enhancement and detection performance of the network without APD. The qualitative results for underwater image enhancement are shown in Fig.~\ref{fig:visualizationforapd}. 
It is clear that when APD is not utilized, the enhancement performance is significantly affected by the attack. The distant background is heavily disturbed by noise, resulting in limited color restoration. In contrast to the unsatisfactory results, the proposed method not only mitigates the interference from the attack to a greater extent but also reduces the undesired blue cast caused by light absorption and scattering. Additionally, we provide color distribution graphs for the RGB channels of each image. It is evident that the added perturbations are visually indistinguishable and have a minimal impact on the distribution. However, the enhanced images with added attacks show noticeable changes in their objective pixel distribution. Compared to the results with APD, our method demonstrates less impact from the attacks.
Fig.~\ref{fig:ablationapd} presents the quantitative results of our method in image enhancement and object detection. (a) and (b) denote the evaluation on enhancement results with unattacked and attacked images. (c) shows the comparison on detection results. The proposed method exceeds the method without APD in all metrics, which demonstrate the superiority of APD.

\section{Conclusion}
This work proposes an precision and robustness collaborative network to enhance underwater images against various attacks, resulting in improved visual and detection predictions. In the forward process, underwater images are transformed into latent components across different frequency domains for isolating degradation factors. Then we reconstruct enhanced images without attacks in the backward process. We also employ an bilevel attack optimization strategy for both visual and perceptual attacks, enabling the network to remove multiple attack types. 
In addition, we introduce the attack pattern discriminator to adaptively adjust the network parameters according to the type of attack on the input image to a ensure reliable robustness.
Experimental results on both clean and attacked images illustrated that the proposed method can achieve robust enhancement and detection results.
%

\bibliographystyle{IEEEtran}
\bibliography{bib2020template}

%
%
%
%
%

\vfill

\end{document}